\documentclass[runningheads]{llncs}

\usepackage{eccv}

\usepackage{eccvabbrv}

\usepackage{graphicx}
\usepackage{booktabs}
\usepackage[title,toc,titletoc,header]{appendix}

\usepackage[accsupp]{axessibility}  
\usepackage[utf8]{inputenc}
\usepackage{multirow}

\usepackage[hidelinks]{hyperref}
\usepackage{orcidlink}

\begin{document}

\title{MagicMakeup: A Region-Controllable Diffusion Transformer for High-Fidelity Makeup-Transfer} 

\titlerunning{MagicMakeup}


\author{Ziyi Wang$^{1,2\star}$\orcidlink{0009-0000-5449-9095},
Siming Zheng$^{2}$\thanks{Equal contributions;$^{\dagger}$ Corresponding author.}\orcidlink{0000-0002-2823-3901},
Yang Yang$^{1,2}$\orcidlink{0009-0000-0622-1275},
Shusong Xu$^{2}$\orcidlink{0009-0009-1829-2826}, \\
Hao Zhang$^{2}$\orcidlink{0009-0007-1175-5918}, 
Bo Li$^{2}$\orcidlink{0000-0001-7817-0665},
Changqing Zou$^{1\dagger}$\orcidlink{0000-0001-8264-6849}, 
Peng-Tao Jiang$^{{2\dagger}}$\orcidlink{0000-0002-1786-4943}
}

\authorrunning{Wang, Zheng, Yang et al.}

\institute{State Key Lab of CAD\&CG, Zhejiang University, China
\and vivo BlueImage Lab, vivo Mobile Communication Co., Ltd., China\\}

\definecolor{first}{RGB}{255,200,200}

\maketitle

\begin{center}
\small
\url{https://vivocameraresearch.github.io/magicmakeup}
\end{center}

\begin{figure}[th]
\centering
\includegraphics[width=1\textwidth]{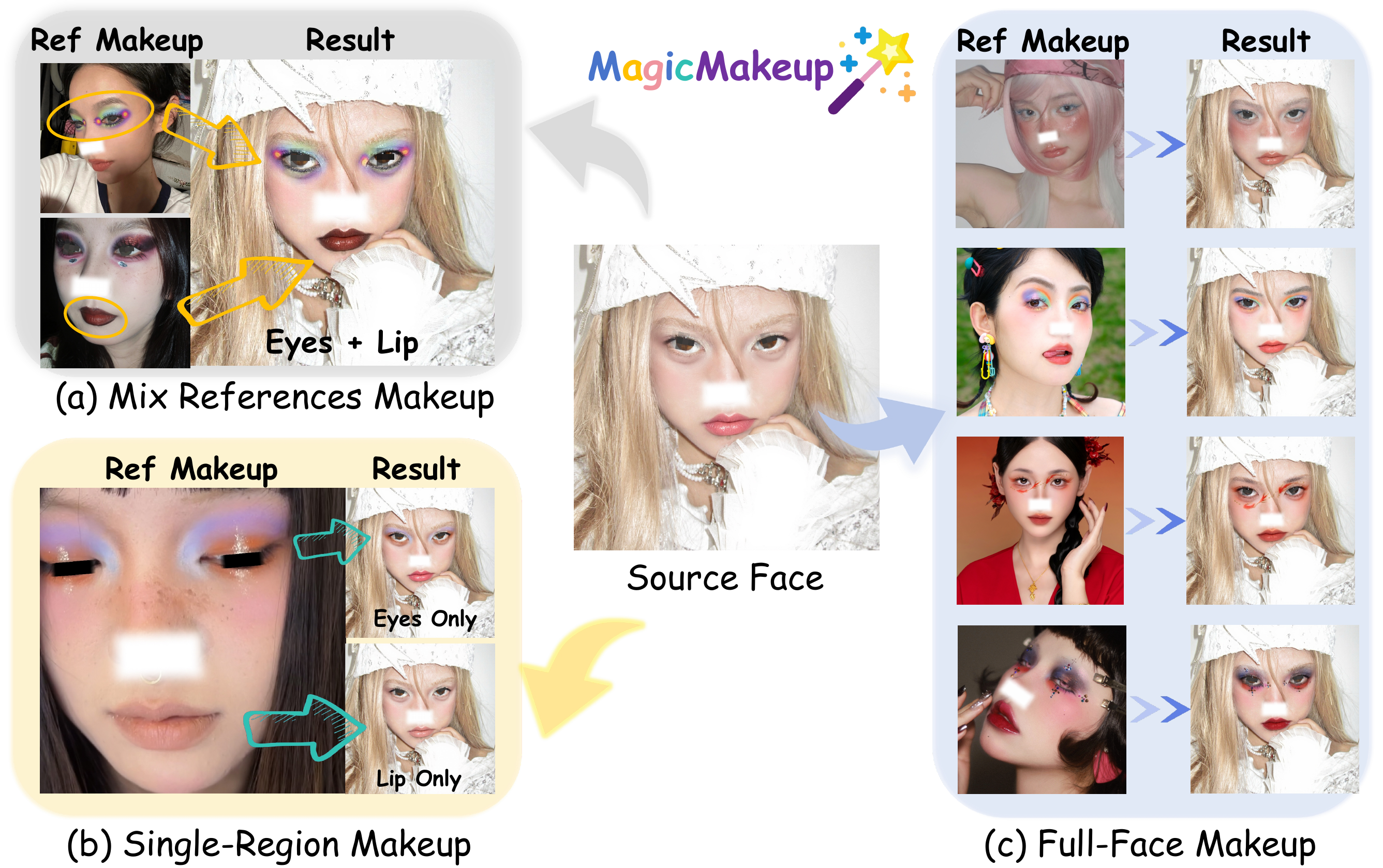}
\caption{\textbf{MagicMakeup: a diffusion transformer for high-fidelity, region-controllable makeup transfer.} It supports mixing references for eyes and lips, single-region editing, and full-face makeup application.}
\label{fig: teaser}
\end{figure}

\begin{abstract}
Makeup-transfer applies the reference makeup to the source face while preserving the source identity. Despite advances in full-face editing by diffusion-based methods, strong regional controllability, makeup fidelity, and identity preservation remain challenging. The reasons are (i) pixel-to-attention misalignment that causes spillover into non-target areas and weakens regional control; (ii) unclear transfer/preservation concept separation under two-image conditioning, leading to coupling between makeup attributes and identity; and (iii) the lack of a high-resolution dataset that is identity-consistent and region-labeled for fine-grained supervision. In this paper, we propose MagicMakeup, a diffusion transformer-based framework for region-controllable and high-fidelity makeup transfer, built on spatial constraints and concept disentanglement. To enable precise region-specific editing while preserving identity, we propose Token-Aligned Region Gating, which aligns pixel masks with attention and applies region-specific logit gating. To clarify the concepts of transfer and preservation, we further introduce Cross-Modal Perception Guidance, which aligns text and image features to enhance cross-modal concept perception. We also design a pipeline for the generation of $1024 \times 1024$ data pairs through region-specific makeup removal and establish a unified benchmark in synthetic and real settings. Extensive quantitative and qualitative experiments show that MagicMakeup improves regional controllability, makeup fidelity, and identity preservation, with strong robustness across styles, races, and poses.
\end{abstract}

\section{Introduction}
\label{sec:intro}
Given a source face and a makeup reference, Makeup-transfer generates an output that preserves the source identity and facial geometry while transferring the reference makeup (color, texture, style), supporting full-face and regional editing. Compared with generic face editing, makeup-transfer places greater emphasis on identity preservation, region-specific control, and realistic high-resolution results, supporting applications such as virtual makeup try-on and personalized digital avatars.

\begin{figure*}[t]
\begin{center}
\includegraphics[width=\linewidth]{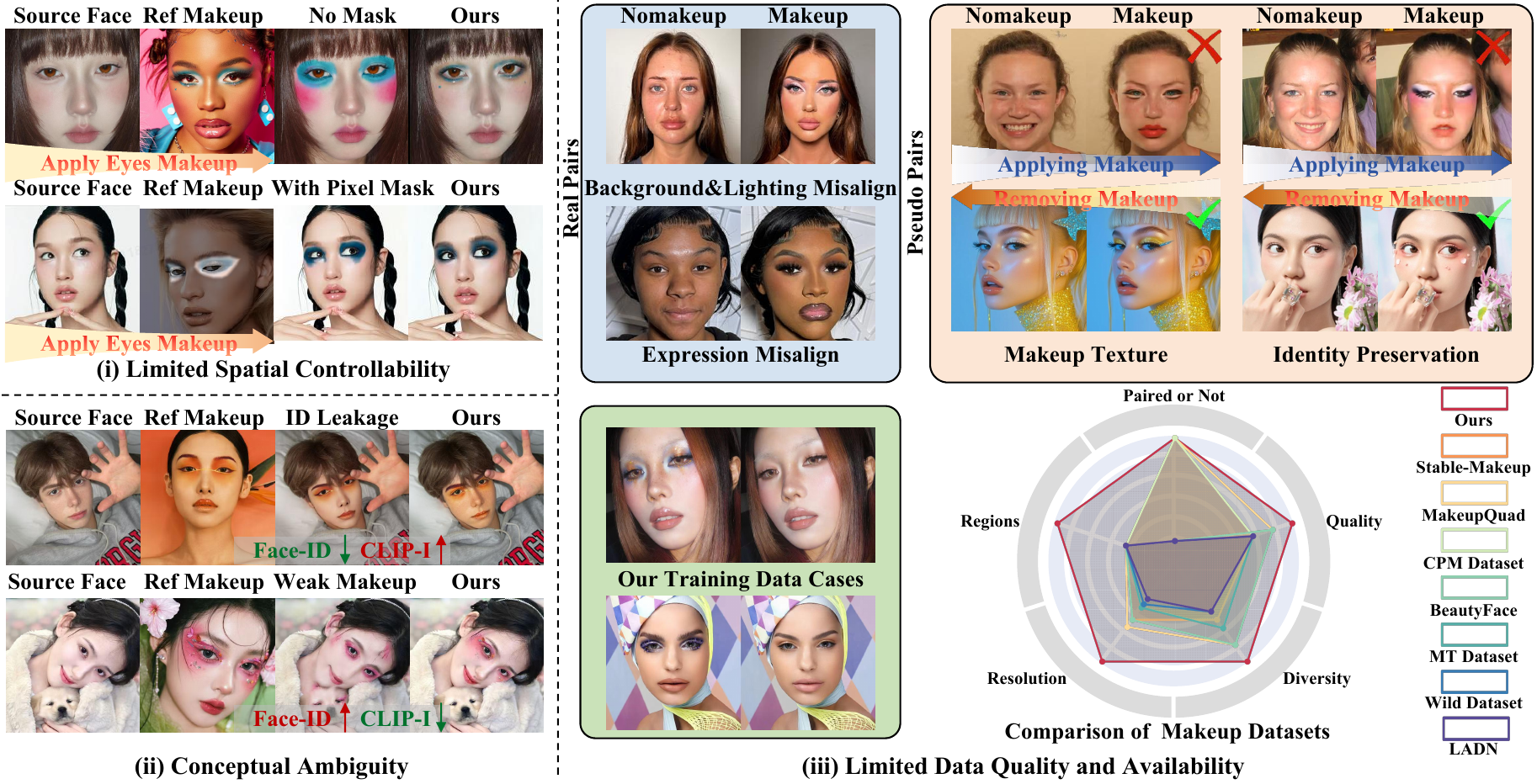}
\end{center}
\caption{\small
\textbf{Three main challenges in makeup transfer:} (i) Limited Spatial Controllability, where pixel-to-attention misalignment prevents reliable region-specific control, causing makeup spillover into non-target areas and weakening precision in localized edits. (ii) Conceptual Ambiguity, where the separation between identity preservation and makeup transfer is unclear, leading to identity–attribute coupling. This causes either the reference facial structure to leak into the output or the makeup details to be inadequately transferred.  (iii) Limited data quality and availability, where real makeup/non-makeup pairs are scarce and often misaligned in expression, background, and lighting, leading to reliance on makeup-applied pseudo-pairs that lack texture, compromise identity, and introduce distribution shift.}
\label{fig: 11}
\end{figure*}

Recently, several methods \cite{sun2024shmt,sun2024diffam,sun2024content,zhang2025stablemakeup,ruan2025mad,zhu2025flux} have been proposed for the makeup-transfer task. Most use pretrained diffusion models or Diffusion Transformers (DiT), integrate reference makeup through cross-attention or feature concatenation, and are trained on unpaired or synthetic pseudo-paired data. However, achieving regional controllability, makeup fidelity, and identity preservation simultaneously remains challenging for three reasons, as shown in \cref{fig: 11}:

\textbf{(i) Limited spatial controllability.} Many diffusion methods \cite{sun2024diffam,sun2024shmt,zhang2025stablemakeup,zhu2025flux} support only full-face transfer. Even with pixel masks \cite{ruan2025mad}, constraints misalign in the attention space, causing spillover and weakening region-specific control and precision.
\textbf{(ii) Conceptual ambiguity.} In two-image conditioning with a source face and a makeup reference, current models often blur the transfer/preservation concepts. They fail to restrict transfer to makeup attributes while preserving source identity and facial geometry, which couples makeup and identity. As a result, the makeup details are not faithfully transferred, or the reference facial structure leaks into the output.
\textbf{(iii) Limited data quality and availability.} High-quality real makeup/non-makeup pairs are scarce and often misaligned in expression, background, and lighting, which makes them unsuitable for supervised training. As a result, most works use unpaired data \cite{li2018beautygan,gu2019ladn,jiang2020psgan,zhao2020scgan,xiang2022ramgan,yang2022elegant,yan2023beautyrec,ruan2025mad} or synthesize pseudo-pairs by applying makeup to the source via text prompts \cite{nguyen2021lipstick,zhang2025stablemakeup,zhu2025flux}, but the generated results lack rich makeup texture and often fail to preserve identity, leading to a distribution shift from real data. 
In addition, existing benchmarks lack region annotations and are of low resolution \cite{li2018beautygan,gu2019ladn,jiang2020psgan,yan2023beautyrec}, which limits fine-grained, region-specific evaluation.

To mitigate these gaps, we propose MagicMakeup, a DiT-based framework for region-controllable, high-fidelity makeup transfer, built on token-aligned region constraints and transfer/preservation disentanglement.
To impose region-specific constraints, we introduce Token-Aligned Region Gating (TARG), which maps pixel masks to tokens and applies region-specific logit gating. To clarify the transfer/preservation concepts, we introduce Cross-Modal Perception Guidance (CMPG), which strengthens cross-modal alignment between text and image features to decide which makeup attributes to transfer from the makeup reference and which identity features to preserve from the source image. 

Furthermore, we develop an automated data pipeline to construct paired makeup and non-makeup images with region masks. We obtain pairs via region-specific makeup removal on $1024 \times 1024$ real makeup images rather than synthesizing makeup. Makeup removal better preserves facial structure, producing non-makeup data closer to the real-world distribution, while the original makeup images retain diverse styles and textures, providing realistic, aligned supervision.
Finally, we establish a unified high-resolution benchmark across synthetic and real settings, covering diverse demographics and fine-grained facial regions, with standardized evaluation of identity preservation, makeup fidelity, and regional control accuracy.
Our contributions to the community are threefold:

\begin{itemize}
    \item We propose MagicMakeup, a DiT-based, region-controllable framework achieving state-of-the-art performance across diverse styles and real-world data and showing strong robustness and generalization.
\end{itemize}

\begin{itemize}
    \item We introduce TARG and CMPG to (i) enforce token-aligned, region-specific attention constraints and (ii) disentangle transfer from preservation, enabling precise regional makeup transfer with reduced spillover and improved identity consistency.
\end{itemize}

\begin{itemize}
    \item We build an automated ``real makeup to regional makeup removal'' data construction pipeline that produces high-resolution, identity-consistent, texture-faithful pairs enabling realistic, well-aligned supervision. And we establish a unified high-resolution benchmark that spans synthetic and real settings for fair evaluation.
\end{itemize}
\section{Related Work}
\label{sec:related}

\subsection{Facial Makeup-Transfer}
Makeup-transfer has advanced rapidly. Early GAN methods \cite{li2018beautygan,chen2019beautyglow,yan2023beautyrec} preserved identity via alignment and identity/makeup disentanglement. PSGAN \cite{jiang2020psgan} captures local detail but needs hand-tuned boundaries; CPM \cite{nguyen2021lipstick} uses UV correspondence yet fails across identities and large style gaps; SCGAN \cite{zhao2020scgan} and EleGANt \cite{yang2022elegant} introduce attention and region-wise semantics. Overall, these methods depend on strict alignment and delicate losses, and still suffer mask misalignment and boundary leakage under large style changes.

With the rise of diffusion models \cite{dhariwal2021diffusion,rombach2022high,hou2024high,podell2023sdxl,ramesh2022hierarchical,zhang20243d,saharia2022photorealistic}, newer methods address content boundaries more directly through stronger generation and semantic conditioning. DiffAM \cite{sun2024diffam} enables attribute-conditioned edits, but sparse or ambiguous labels make the preserve/transfer boundary unstable. SHMT \cite{sun2024shmt} builds pseudo-paired data via a 3D face model with Laplacian contours and stabilizes geometry using a hierarchical Transformer, but it still suffers from boundary noise and weak fusion in overlapping regions. Stable-Makeup \cite{zhang2025stablemakeup} uses CLIP-guided text and image conditioning with face-control modules, but once alignment drifts, errors amplify and spread across regions. MAD \cite{ruan2025mad} enables region-wise makeup application but still suffers from makeup spillover, concept ambiguity, and high computational cost at high resolutions. Despite progress, the field still lacks a testable criterion for deciding what to preserve and what to transfer, and a spatial control mechanism fully aligned with internal representations.

\subsection{Makeup-Transfer Datasets}
Given the difficulty of collecting high-quality paired makeup data, makeup transfer has long relied on unpaired training. MT \cite{li2018beautygan} introduced this paradigm with surrogate losses such as color-histogram matching. LADN \cite{gu2019ladn} expanded data diversity and keypoint annotations, and Makeup-Wild \cite{jiang2020psgan} further increased scale, but source–reference identity mismatch limited strict supervision. MT-HR \cite{liu2021psgan++}, and BeautyFace \cite{yan2023beautyrec} improved resolution and annotations, and MT-Text \cite{ruan2025mad} added text-driven control. However, strongly paired makeup/non-makeup samples with the same identity, similar pose, and clear region masks remain rare, and available data are often low resolution (e.g., 256×256). These gaps hinder fine-grained, controllable, and distribution-faithful supervision.

With the rise of image editing models \cite{bodur2024iedit,chefer2023attend,zhang2023sine,cao2023masactrl,mou2023dragondiffusion,li2023layerdiffusion,huang2024smartedit,wu2025qwen,zhang2025context,huang2025diffusion,labs2025flux,feng2025dit4edit,azizi2023synthetic}, many works turn to pseudo-paired data: SHMT \cite{sun2024shmt} derives pairs from a single reference using a 3D face model \cite{guo2020towards} with Laplacian contours; Stable-Makeup \cite{zhang2025stablemakeup}, BeautyBank \cite{lu2025beautybank} and Flux-Makeup \cite{zhu2025flux} apply makeup to source images to synthesize pairs at scale, but limited application quality and predefined styles yield makeup outputs with limited identity alignment, style diversity, and visual quality, causing distribution shift from real data. Moreover, support for region-wise editing is weak. Thus, building high-quality, style-diverse, and scalable strongly paired datasets is a key challenge for makeup-transfer.
\section{Method}
\label{sec:method}

\subsection{Overview}

\begin{figure*}[t]
\begin{center}
\includegraphics[width=\linewidth]{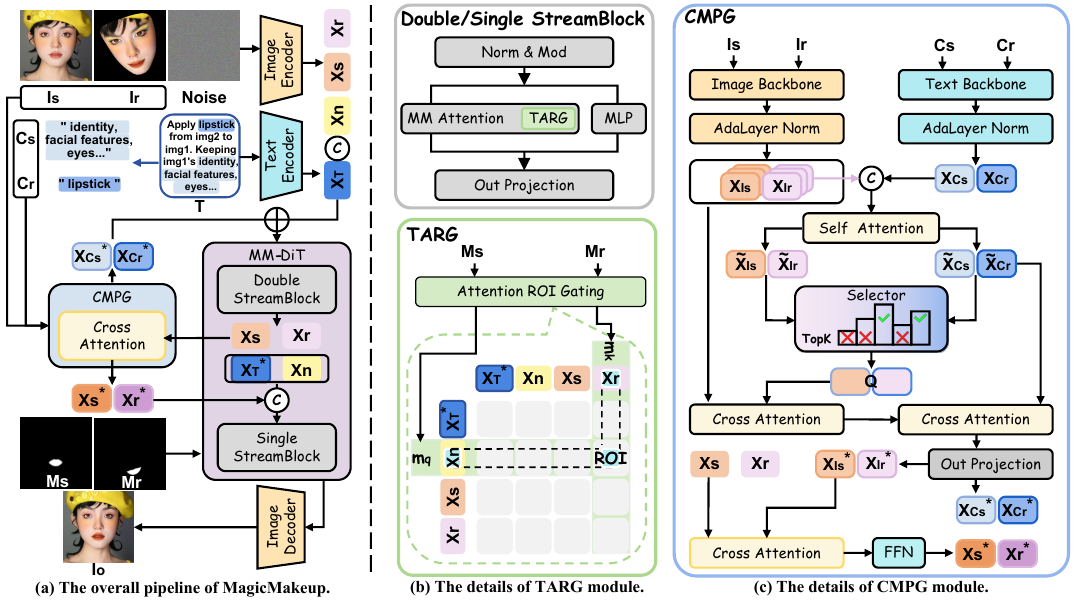}
\end{center}
\caption{\small
\textbf{The overview of our MagicMakeup pipeline, as shown in (a).} We introduce Token-Aligned Region Gating within DiT’s multi-modal attention, which maps pixel masks to tokens and applies attention ROI gating to impose region-specific constraints, as shown in (b). We introduce Cross-Modal Perception Guidance, which strengthens cross-modal alignment between text and image features to clarify the preservation/transfer concepts, as shown in (c). }
\label{fig: pipeline}
\end{figure*}

The overall pipeline of MagicMakeup is illustrated in \cref{fig: pipeline}. Given a source face \( \mathbf{I}_s \) and a reference makeup image \( \mathbf{I}_r \), the goal is to generate an output image \( \mathbf{I}_o \) that transfers the fine-grained makeup details from \( \mathbf{I}_r \) while preserving the identity and facial structure of \( \mathbf{I}_s \).
To achieve this, we first encode both \( \mathbf{I}_s \) and \( \mathbf{I}_r \) into token sequences \( \mathbf{X}_s \) and \( \mathbf{X}_r \), and sample a noise token \( \mathbf{X}_n \) for the generation process. Simultaneously, the text prompt \( \mathbf{T} \) is encoded into text embeddings \( \mathbf{X}_T \). These visual and textual tokens are then processed jointly by our framework to iteratively denoise the latent token \( \mathbf{X}_n \) and synthesize the target image tokens.

To ensure makeup is transferred only to designated regions of the face and to prevent spillover, we introduce TARG, which uses region masks \( \mathbf{M}_s \) and \( \mathbf{M}_r \) for the source and reference faces, respectively, to restrict attention within the target facial regions. 
Additionally, to clarify and strengthen the concepts of preservation and transfer, we propose CMPG, which modulates both the image and text features. 
CMPG aligns the preservation concept \( \mathbf{C}_s \) with the source image \( \mathbf{I}_s \) and the transfer concept \( \mathbf{C}_r \) with the reference image \( \mathbf{I}_r \), updating both image and text features accordingly. The updated features \( \mathbf{X}_s^*, \mathbf{X}_r^*, \mathbf{X}_{C_s}^* \), and \( \mathbf{X}_{C_r}^* \) are then integrated by MM-DiT.
Finally, an image decoder reconstructs the denoised tokens into the final output image \( \mathbf{I}_o \), successfully transferring the makeup of \( \mathbf{I}_r \) and retaining the identity of \( \mathbf{I}_s \).

\subsection{Token-Aligned Region Gating}

Most existing diffusion-based makeup-transfer pipelines are optimized for full-face transfer, making region-specific edits on eyes and lips unreliable. To enable precise transfer on three target regions, namely face, eyes, and lips, we propose Token-Aligned Region Gating (TARG) using refined facial masks. A key challenge is that pixel masks do not reliably constrain attention-based information flow, since pixel masks are misaligned with token interactions, allowing makeup cues from the reference to leak outside the region of interest (ROI). TARG resolves this mismatch by projecting the ROI mask onto the token grid and enforcing the constraint directly in attention.

For each target region (face, eyes, or lips), we downsample and flatten the source and reference masks
$\mathbf{M}_s$ and $\mathbf{M}_r$ to the token grids of MM-Attention, yielding a latent query gate
$\mathbf{m}_q\in\{0,1\}^{L_{\text{n}}}$ and a reference key gate $\mathbf{m}_k\in\{0,1\}^{L_r}$, with lengths $L_{\text{n}}$ and $L_r$ for $\mathbf{X}_n$ and $\mathbf{I}_r$.
We define an additive logit mask on latent-to-reference token pairs as
\begin{equation}
M_{i,j}=
\begin{cases}
0, & \mathbf{m}_q(i)=1 \ \text{and}\ \mathbf{m}_k(j)=1,\\
-\infty, & \text{otherwise},
\end{cases}
\end{equation}
where $i$ and $j$ index latent-query and reference-key tokens, respectively.
In practice, $-\infty$ is implemented with a sufficiently large negative constant.
The masked attention is computed as
\begin{equation}
\mathrm{Attn}(\mathbf{Q},\mathbf{K},\mathbf{V})
=\operatorname{Softmax}\!\left(\frac{\mathbf{Q}\mathbf{K}^\top}{\sqrt{d_k}}+\mathbf{M}\right)\mathbf{V}.
\end{equation}
We apply $\mathbf{M}$ only to the latent-query and reference-key interactions and keep all other attention terms unchanged.
As a result, latent tokens outside the ROI cannot attend to $\mathbf{I}_r$, suppressing cross-region leakage and enabling precise region-level makeup transfer.

\subsection{Cross-Modal Perception Guidance}

Unlike generic face editing, makeup-transfer requires conditioning on two images: the source face and the makeup reference. This makes it challenging to transfer makeup attributes while preserving identity geometry without mixing them.
To address this, we propose Cross-Modal Perception Guidance (CMPG), which semantically decouples what to preserve and what to transfer, and jointly strengthens both the image and text representations for these two concepts.

From the instruction $\mathbf{T}$, we extract two concept phrases: a preservation concept $\mathbf{C_s}$ and a transfer concept $\mathbf{C_r}$. Intuitively, $\mathbf{C_s}$ describes identity-related cues (e.g., \textit{identity, facial features...}), while $\mathbf{C_r}$ specifies the target makeup attributes in the edited region (e.g., \textit{lipstick} for lip makeup or \textit{eyeshadow} for eye makeup).
At each denoising step (Fig.~\ref{fig: pipeline}(c)), CMPG processes two concept-image pairs $(\mathbf{C_s},\mathbf{I_s})$ and $(\mathbf{C_r},\mathbf{I_r})$ and outputs concept-aware text modulations $\mathbf{X}_{C_s}^*, \mathbf{X}_{C_r}^*$ together with image modulations $\mathbf{X}_{I_s}^*, \mathbf{X}_{I_r}^*$. The goal is to make preservation/transfer cues more explicit in the text stream and to make each image stream respond to its corresponding concept without cross-contamination.

\subsubsection{Concept-guided evidence selection.} We extract concept tokens $\mathbf{X}_C$ from a frozen text backbone and image tokens $\mathbf{X}_I$ from the last layers of a frozen image backbone, where $(\mathbf{C},\mathbf{I})$ denotes the current pair, i.e., $(\mathbf{C_s},\mathbf{I_s})$ or $(\mathbf{C_r},\mathbf{I_r})$. For each concept-image pair, we first co-update the two token sets with a joint self-attention block to align concept semantics with responsive image regions, yielding $\tilde{\mathbf{X}}_C$ and $\tilde{\mathbf{X}}_I$. 
To focus subsequent updates on concept-relevant evidence, we rank image tokens by their similarity to the concept tokens and select a compact set of top-$k$ visual queries:
\begin{equation}
\mathbf{Q}=\tilde{\mathbf{X}}_{I}\Big[\mathrm{Top}\text{-}k\big(\mathrm{Max}(\tilde{\mathbf{X}}_{I}\tilde{\mathbf{X}}_{C}^{\top})\big)\Big],
\end{equation}
where $\mathrm{Max}(\cdot)$ pools similarities over concept tokens. $\mathbf{Q}$ captures the most concept-aligned regions and serves as shared evidence for updating both modalities.

\subsubsection{Updating image and text perceptions.} Starting from $\mathbf{Q}$, we refine the queries with two cross-attentions. The first attends to a multi-layer visual context bank (from the last three backbone layers) of the corresponding image, and the second attends to the concept tokens. AdaLayerNorm heads map the refined queries to image modulation and span-aware text modulation.

We write the text modulation $\mathbf{X}_{C_s}^*, \mathbf{X}_{C_r}^*$ only on the token spans of $\mathbf{C_s}$ and $\mathbf{C_r}$ in the text sequence $\mathbf{X}_T$, which is then fed into MM-DiT as textual conditioning. 
The image modulation  $\mathbf{X}_{I_s}^*, \mathbf{X}_{I_r}^*$ is injected into the corresponding image streams $\mathbf{X}_s$,$\mathbf{X}_r$ via a cross-attention adapter, and the modulated features are further processed by the DiT single-stream blocks.

By strengthening concept-aware representations for both text and images at each denoising step, CMPG provides timestep-conditioned and concept-disentangled guidance, reducing identity leakage and attribute drift while improving local makeup fidelity.

\begin{figure*}[t]
\begin{center}
\includegraphics[width=\linewidth]{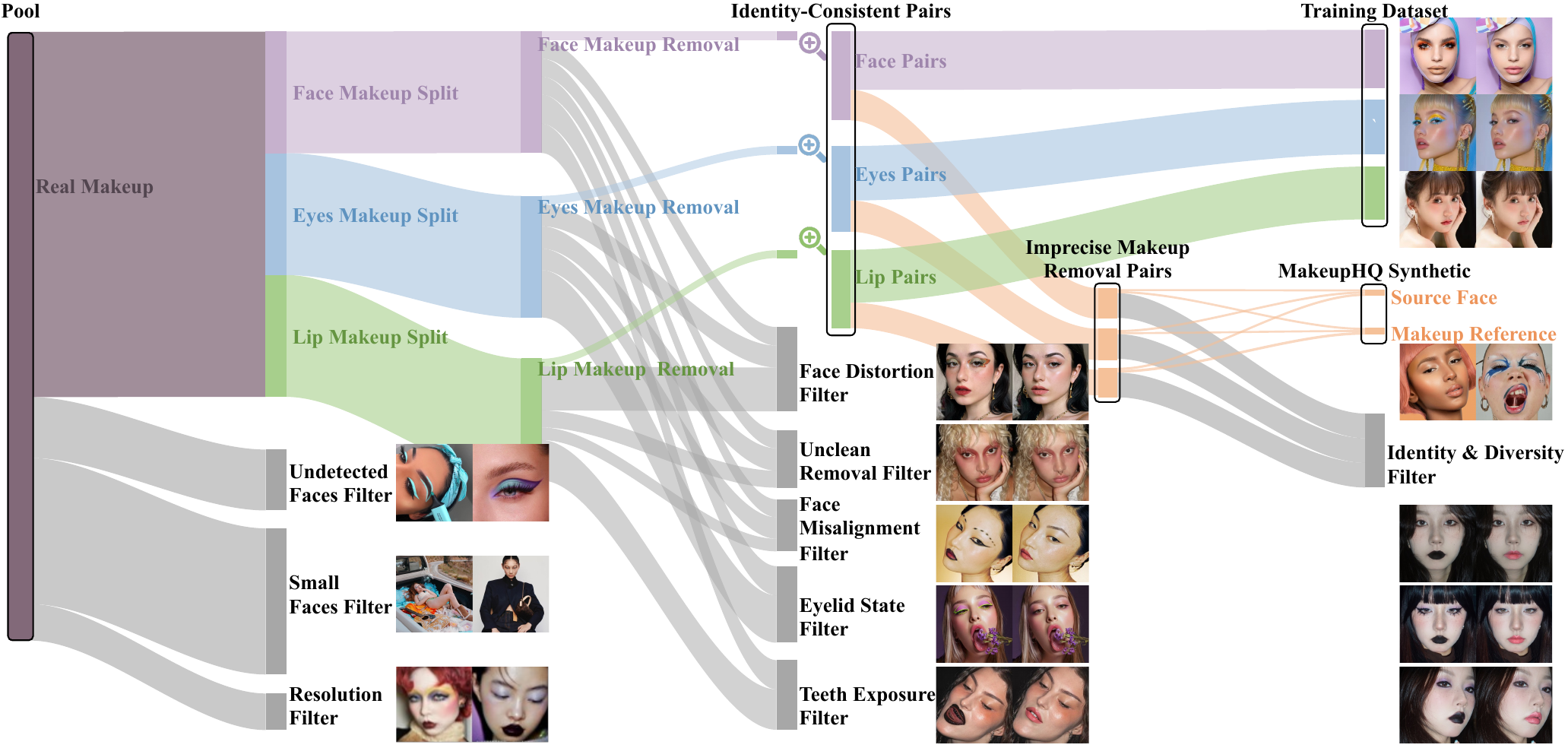}
\end{center}
\caption{\small
\textbf{Diagram of our training and synthetic dataset construction pipeline.} It shows the filtering stages for face distortion, unclean removal, and imprecise regions, resulting in identity-consistent, region-labeled makeup/non-makeup pairs across face, eye, and lip splits.}
\label{fig: datapipe}
\end{figure*}

\subsection{Dataset Construction Pipeline}

\subsubsection{Makeup/No-Makeup Pairs.} In practice, we attempted to synthesize paired data by applying makeup with Flux-Kontext \cite{labs2025flux}, GPT-4o \cite{hurst2024gpt}, and Nano-Banana \cite{team2023gemini}. Both text-guided and reference-image-guided generation are unreliable: the former often yields unrealistic makeup with weak details and limited diversity, while the latter frequently distorts facial structure and still fails to faithfully transfer the reference makeup as shown in \cref{fig: 11} (iii).
We therefore construct pairs via region-specific makeup removal from real makeup photos. This is more stable because facial geometry is strongly constrained by the input and the model’s facial prior, whereas adding makeup requires inventing high-frequency, style-specific textures. Makeup removal acts like subtracting a local appearance layer, preserving identity and providing realistic, aligned supervision.

As illustrated in \cref{fig: datapipe}, we begin with a pool of 50,000 real $1024 \times 1024$ makeup faces spanning diverse styles and poses. After removing undetected, very small faces, and low-resolution samples, we crop and normalize the remaining images. This produces about 3,000 standardized images, which are then grouped into face, eyes, and lip makeup splits. For each split, FLUX-Kontext \cite{labs2025flux} performs region-specific makeup removal (full face, eyes, lips), generating multiple non-makeup candidates which are then matched with corresponding makeup pairs.

We then apply a cascade of filters. First, identity-consistent pairs are obtained by removing face distortion and misalignment: DFA-MobileNet \cite{kim2024keypoint} aligns faces, AdaFace \cite{kim2022adaface} similarity removes unchanged or distorted results, and landmark-mask differences eliminate eyelid-state and teeth-exposure mismatches. Next, an unclean-removal filter ensures sufficient regional color-histogram change; among remaining candidates, the best pair per group is selected using a balanced score combining identity similarity and regional color difference. This yields around 10,000 clean, identity-consistent pairs. Finally, an imprecise-region filter enforces large change within the labeled region and small change elsewhere, leaving 6,772 region-labeled pairs (2,392 face, 2,224 eyes, 2,156 lip). During training, the non-makeup image serves as the source and the makeup image as the GT. 

Compared with applying makeup to bare faces, removal better preserves facial structure for the non-makeup, while the makeup retains authentic texture and style diversity. This provides realistic and aligned supervision for training.

\subsubsection{Reference Makeup.} A challenge is to construct references that differ in identity from the makeup image while preserving the makeup appearance. Prior method \cite{zhang2025stablemakeup} modifies only a small subset of facial landmarks, yielding insufficient deformation and causing structural leakage. We instead construct an identity-separated reference by choosing, for each makeup image, a face with a large identity difference as the deformation target and warping makeup to that geometry using 478 facial landmarks, Delaunay triangulation with piecewise-affine transforms, and edge smoothing. The resulting reference image preserves the makeup image’s makeup texture and style but adopts a new identity and facial layout, which better decouples identity from makeup.

\subsubsection{Masks.}
We obtain a full-face mask from face parsing \cite{xie2021segformer} and generate eye and lip masks from the MediaPipe Face Landmarker (468 points) \cite{kartynnik2019real} using predefined region landmark sets. Since eyeshadow is not directly available, we construct a soft eyeshadow mask by dilating and shifting the eye mask, multiplying it with a radial gradient, and smoothing the boundary.

\subsection{MakeupHQ Benchmark}
Current benchmarks are low-resolution, lack dedicated splits for full-face, eye, and lip makeup, and exhibit small identity gaps with limited style, pose, and expression diversity, which fails to support the fine-grained, region-specific makeup-transfer needed in real-world applications.
To address this, we introduce MakeupHQ Bench as a supplementary benchmark for makeup transfer, consisting of two test sets: (i) MakeupHQ-Synthetic, a synthetic cross-identity split selected from the remaining pool by identity and diversity filtering, and (ii) MakeupHQ-Real, an independent real-world split with real non-makeup sources and real makeup references. MakeupHQ-Real is image- and identity-disjoint from both the training data and MakeupHQ-Synthetic.

All source images and makeup references are standardized using the same preprocessing pipeline, including detection, center-cropping, normalization, parsing, and landmark-based mask extraction, followed by careful manual screening. MakeupHQ-Synthetic contains 267 makeup references, including 139 eye references and 128 face or lip references, together with 277 source images. MakeupHQ-Real contains 295 makeup references, including 174 eye references and 121 face or lip references, together with 243 source images. 
Style coverage includes Western, Japanese/Korean, East Asian, and neo-Chinese; poses range from frontal to profile across multiple viewpoints; expressions include open/closed eyes and mouth, with or without teeth. Compared with prior datasets, MakeupHQ Bench offers higher resolution, broader style diversity, clearer region labels, and a distribution closer to real data, enabling fine-grained and controllable evaluation.
\section{Experiments}
\label{sec: exp}

\subsection{Datasets and Metrics}

We evaluate MagicMakeup on our benchmark, which includes two high-resolution datasets at $1024 \times 1024$: MakeupHQ-Synthetic (500 pairs) and MakeupHQ-Real (500 pairs). For comparability with prior work, we additionally report performance on the public Makeup-Wild dataset \cite{jiang2020psgan} (1000 pairs) with diverse poses at $256 \times 256$.

We assess four aspects of quality: identity fidelity, makeup similarity, unedited-region consistency, and distribution quality. For identity fidelity, we extract AdaFace embeddings \cite{kim2022adaface} after standard face alignment and compute cosine similarity, which offers stronger discriminativeness and makeup invariance than SSIM. Following Stable-Makeup \cite{zhang2025stablemakeup}, we use CLIP-I \cite{radford2021learning} and DINO-I \cite{caron2021emerging} to measure cosine similarity between the makeup reference and the generated image. Unedited-region consistency is evaluated using L2-M, which computes the mean squared error on masked non-edited regions (background for full-face transfer) between the source and generated images. Finally, we report FID \cite{heusel2017gans} between generated results and real makeup images in each test split using Inception features to assess overall distribution quality.

\subsection{Implementation Details}

We train our MagicMakeup on our training set using 8 NVIDIA A100 80GB GPUs with a batch size of 2 per GPU. We optimize with AdamW at a learning rate of 5e-5 for 36K iterations. During training, the text prompt is restricted to three predefined classes and the LoRA rank is set to 256. At inference time, we use 28 sampling steps with a guidance scale of 2.5 to generate the final results.

\begin{figure}[th]
\centering
\includegraphics[width=1\textwidth]{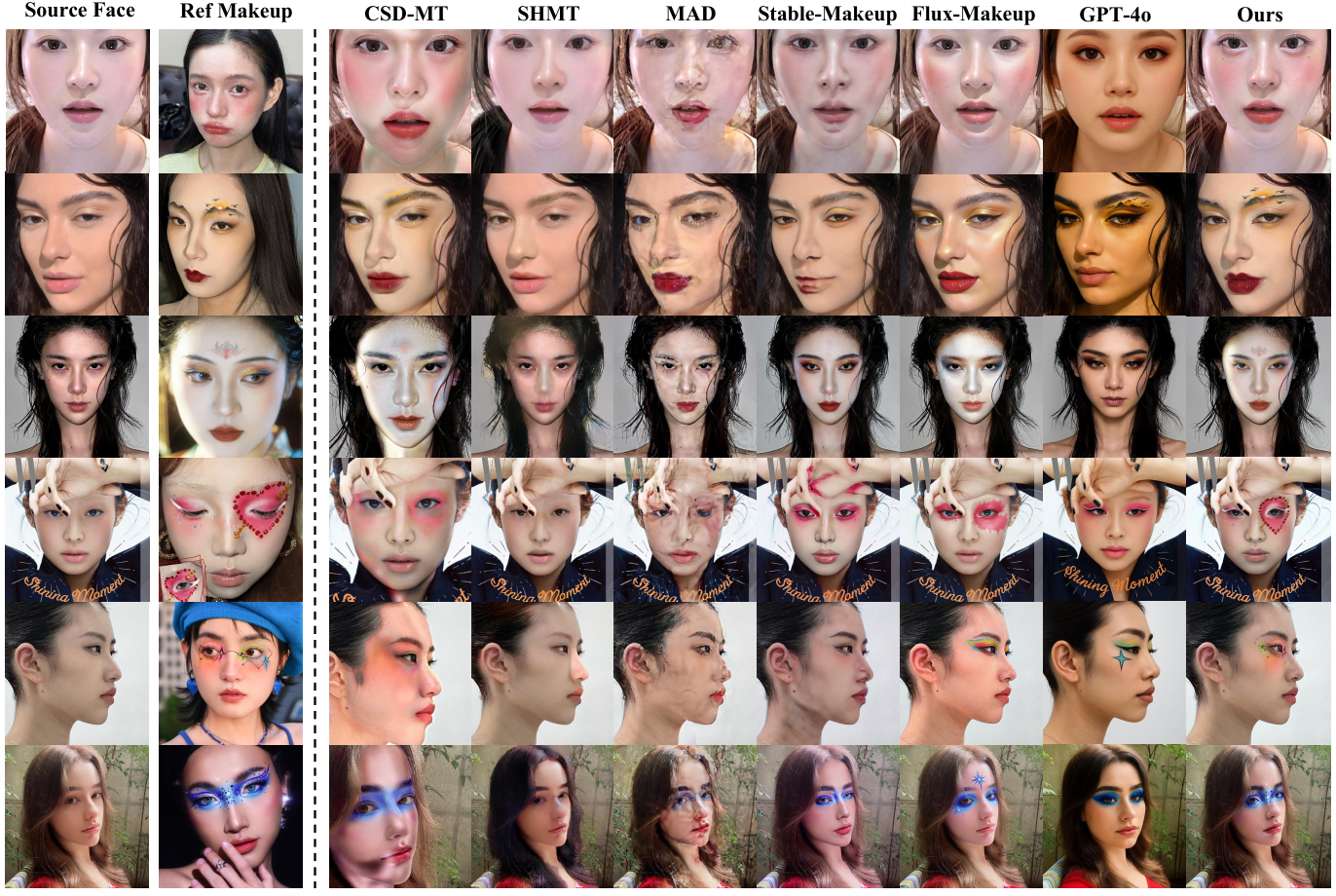}
\caption{\textbf{Qualitative comparison of makeup-transfer results under synthetic and real settings, including diverse poses, backgrounds and makeup styles.} Our method demonstrates superior identity preservation and makeup fidelity. Please zoom in for better visualization.}
\label{fig: Qua1}
\end{figure}

\begin{figure*}[t]
\begin{center}
\includegraphics[width=\linewidth]{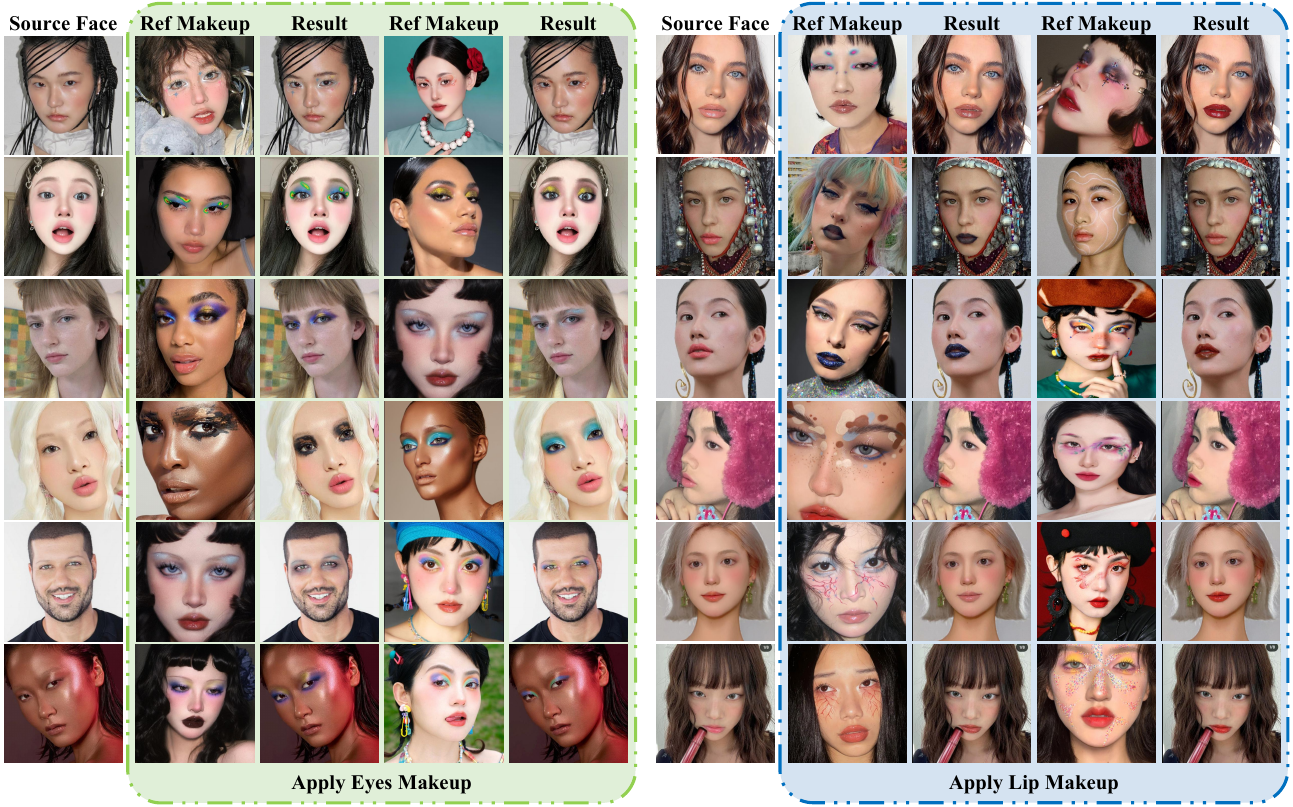}
\end{center}
\caption{\small
\textbf{Demonstration of eye and lip makeup transfer.} Our method applies makeup precisely within the specified regions, producing clean boundaries and fine-grained cosmetic details. The results remain consistent under pose and expression changes and across diverse skin tones, while preserving non-edited facial areas and keeping the background free of spillover or geometric distortion.}
\label{fig: Q2}
\end{figure*}

\begin{table*}[t]
\centering
\caption{\textbf{Quantitative comparison of different methods.} Results are shown on MakeupHQ-Synthetic and MakeupHQ-Real at $1024 \times 1024$ resolution, and Makeup-Wild at $256 \times 256$. Arrows indicate the preferred direction for each metric, and red values highlight the best scores.}
\setlength{\tabcolsep}{3pt}
\footnotesize
\begingroup
\renewcommand{\arraystretch}{1.3}%
\resizebox{\textwidth}{!}{%
\begin{tabular}{p{2.0cm}|l|rrrrrrrrrrrrrrr} 
\toprule
\centering \textbf{Dataset} & \centering \textbf{Metric} & \parbox{1.0cm}{\centering \textbf{CPM\\ \cite{nguyen2021lipstick}}} & \parbox{1.3cm}{\centering \textbf{EleGANt\\ \cite{yang2022elegant}}} & \parbox{1.1cm}{\centering \textbf{SSAT\\ \cite{sun2022ssat}}} & \parbox{1.3cm}{\centering \textbf{CSD-MT \cite{sun2024content}}} & \parbox{1.1cm}{\centering \textbf{SHMT\\ \cite{ramesh2022hierarchical}}} & \parbox{1.0cm}{\centering \textbf{MAD\\ \cite{ruan2025mad}}} & \parbox{1.2cm}{\centering \textbf{Stable-\\Makeup \cite{zhang2025stablemakeup}}} & \parbox{1.2cm}{\centering \textbf{Flux-\\Makeup \cite{zhu2025flux}}} & \textbf{Ours} \\
\midrule
\multirow{5}{*}{\parbox{2.0cm}{\centering \textbf{MakeupHQ-Synthetic}\\($1024 \times 1024$)}} 
& \textbf{DINO-I}$\uparrow$ & 0.625 & 0.520 & 0.461 & 0.576 & 0.488 & 0.500 & 0.612 & 0.576 & \colorbox{first}{0.660} \\
& \textbf{CLIP-I}$\uparrow$ & 0.777 & 0.744 & 0.698 & 0.796 & 0.710 & 0.727 & 0.789 & 0.765 & \colorbox{first}{0.816} \\
& \textbf{Face-ID}$\uparrow$ & 0.400 & 0.796 & 0.684 & 0.563 & \colorbox{first}{0.824} & 0.590 & 0.569 & 0.806 & 0.730 \\
& \textbf{L2M}$\downarrow$ & 1610.938 & 2148.089 & 2816.613 & 3334.562 & 823.657 & 180.674 & 101.416 & 111.476 & \colorbox{first}{56.373} \\
& \textbf{FID}$\downarrow$ & 72.191 & 86.841 & 120.193 & 83.896 & 92.601 & 91.372 & 69.733 & 64.119 & \colorbox{first}{59.238} \\
\midrule
\multirow{5}{*}{\parbox{2.0cm}{\centering \textbf{MakeupHQ-Real}\\($1024 \times 1024$)}} 
& \textbf{DINO-I}$\uparrow$ & 0.575 & 0.534 & 0.448 & 0.596 & 0.482 & 0.476 & 0.616 & 0.552 & \colorbox{first}{0.623} \\
& \textbf{CLIP-I}$\uparrow$ & 0.722 & 0.733 & 0.665 & 0.772 & 0.682 & 0.703 & 0.774 & 0.745 & \colorbox{first}{0.799} \\
& \textbf{Face-ID}$\uparrow$ & 0.361 & 0.755 & 0.641 & 0.522 & 0.777 & 0.539 & 0.525 & \colorbox{first}{0.792} & 0.706 \\
& \textbf{L2M}$\downarrow$ & 2158.131 & 4038.093 & 3138.460 & 5572.246 & 593.029 & 183.091 & 163.979 & 104.464 & \colorbox{first}{49.768} \\
& \textbf{FID}$\downarrow$ & 99.105 & 97.251 & 131.545 & 91.931 & 115.319 & 119.164 & 81.342 & 85.427 & \colorbox{first}{76.273} \\
\midrule
\multirow{5}{*}{\parbox{2.0cm}{\centering \textbf{Makeup-Wild}\\(256×256)}} 
& \textbf{DINO-I}$\uparrow$ & 0.608 & 0.587 & 0.569 & 0.602 & 0.564 & 0.545 & \colorbox{first}{0.619} & 0.568 & 0.581 \\
& \textbf{CLIP-I}$\uparrow$ & 0.621 & 0.640 & 0.581 & 0.640 & 0.586 & 0.609 & 0.637 & 0.621 & \colorbox{first}{0.665} \\
& \textbf{Face-ID}$\uparrow$ & 0.406 & 0.777 & 0.706 & 0.499 & 0.847 & 0.598 & 0.721 & \colorbox{first}{0.928} & 0.884 \\
& \textbf{L2M}$\downarrow$ & 605.857 & 1540.744 & 2709.617 & 2151.729 & 1246.029 & 169.281 & 71.525 & 42.609 & \colorbox{first}{25.839} \\
& \textbf{FID}$\downarrow$ & 86.588 & 94.202 & 109.184 & 94.705 & 103.265 & 105.794 & \colorbox{first}{85.132} & 88.708 & 92.924 \\
\bottomrule
\end{tabular}%
}
\endgroup
\label{tab: exp}
\end{table*}

\subsection{Comparison with SOTA Methods}

We compare with representative makeup-transfer approaches, including GAN-based CPM \cite{nguyen2021lipstick}, EleGANt \cite{yang2022elegant}, SSAT \cite{sun2022ssat}, CSD-MT \cite{sun2024content}, and diffusion-based SHMT \cite{sun2024shmt}, MAD \cite{ruan2025mad}, Stable-Makeup \cite{zhang2025stablemakeup} and Flux-Makeup \cite{zhu2025flux}. These methods serve as our baselines for quantitative and qualitative comparison, and all models are evaluated under the same settings on unified benchmarks.

\subsubsection{Qualitative Comparison}

\cref{fig: Qua1} shows side-by-side results on MakeupHQ-Synthetic and MakeupHQ-Real across varied poses, backgrounds, and a spectrum from light to heavy makeup. Prior methods often degrade identity consistency or bleed color outside facial regions, and we observe face rescaling and background stretching in several cases. Their transfers capture only coarse color cues and often miss fine details and full texture, resulting in incomplete makeup. In contrast, MagicMakeup preserves identity and scene geometry while delivering complete and well-aligned transfers. Light styles look clean and natural, and heavy styles retain the reference color distributions and texture details at the correct facial locations. Non-edited regions, including hair and background, remain intact without warping or deformation. 

Beyond full-face transfer, we also demonstrate eye and lip makeup application in \cref{fig: Q2}. The results remain stable under noticeable pose and expression changes, and show consistent, fine-grained placement confined to the target region. Importantly, non-edited areas preserve their original structure and appearance, with no visible spillover or geometric distortion.

\begin{table}[t]
\centering
\caption{\textbf{Quantitative results of ablation study.} The table shows the performance of different variants \textit{w/o} TARG and CMPG.}
\begin{tabular}{lccccccc} 
\toprule
\textbf{Region} & \textbf{Variant} & \textbf{TARG} & \textbf{CMPG} & \textbf{Face-ID$\uparrow$} & \textbf{DINO-I$\uparrow$} & \textbf{CLIP-I$\uparrow$} \\
\midrule
\multirow{3}{*}{Eyes} & 1 & $\times$ & $\times$ & 0.870 & 0.791 & 0.883 \\
    & 2 & $\checkmark$ & $\times$     & 0.890 & 0.799 & 0.890 \\
    & 3 & $\checkmark$ & $\checkmark$ & 0.922 & 0.801 & 0.904 \\
\midrule
\multirow{3}{*}{Lip} & 1 & $\times$ & $\times$ & 0.879 & 0.653 & 0.877 \\
    & 2 & $\checkmark$ & $\times$ & 0.885 & 0.659 & 0.870 \\
    & 3 & $\checkmark$ & $\checkmark$ & 0.971 & 0.673 & 0.884 \\
\midrule
\multirow{3}{*}{Face} & 1 & $\times$ & $\times$ & 0.805 & 0.733 & 0.833 \\
    & 2 & $\checkmark$ & $\times$ & 0.857 & 0.712 & 0.811 \\
    & 3 & $\checkmark$ & $\checkmark$ & 0.858 & 0.753 & 0.840 \\
\bottomrule
\end{tabular}
\label{tab: ablation}
\end{table}

\subsubsection{Quantitative Comparison}

As shown in \cref{tab: exp}, on the two $1024 \times 1024$ datasets MakeupHQ-Synthetic and MakeupHQ-Real, MagicMakeup achieves the highest DINO-I and CLIP-I and the lowest L2M and FID, while maintaining competitive Face-ID. These results indicate better transfer fidelity, image quality, and identity stability than prior methods. Since L2M is computed only on non-edited background regions, our much lower values further confirm that MagicMakeup preserves global geometry and avoids the face rescaling and background stretching observed in other approaches.
On the public 256×256 Makeup-Wild dataset with diverse poses, MagicMakeup also leads in CLIP-I and L2M, and its FID is close to the best baseline, which is consistent with distribution differences at lower resolution rather than a transfer failure.

We note that faithful makeup transfer can lower Face-ID since makeup changes affect identity-related cues, while methods that transfer less makeup may report higher Face-ID. Even so, MagicMakeup remains competitive on Face-ID while improving fidelity and preserving spatial stability across both high-resolution and in-the-wild settings, validating a better balance between transfer strength and identity preservation.

\begin{figure}[t]
\begin{center}
\includegraphics[width=0.8\linewidth]{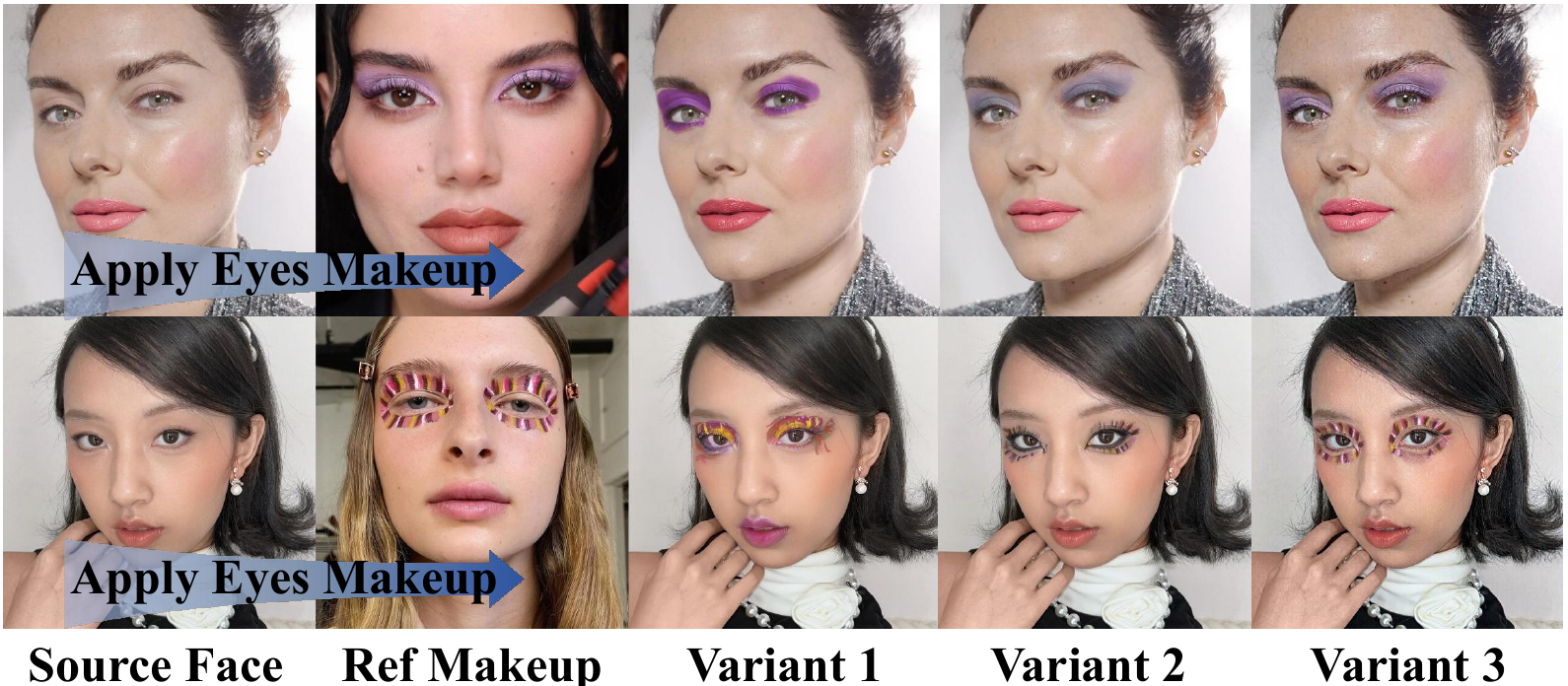}
\end{center}
\caption{\small
\textbf{Qualitative results of ablation study.} Variant 2 with TARG improves region-specific accuracy, while Variant 3 with CMPG further enhances makeup fidelity.}
\label{fig: abla}
\end{figure}

\subsection{Ablation Study}
To evaluate the contribution of each component, we conduct ablation studies on TARG and CMPG. All variants are trained for 10K iterations on our full dataset and tested on the MakeupHQ-Synthetic dataset. Quantitative results are shown in \cref{tab: ablation}, and qualitative results in \cref{fig: abla}.

Starting with the base DiT model, adding Token-Aligned Region Gating (TARG) improves spatial precision: attention is concentrated on the intended regions, and cross-region leakage is effectively suppressed. As shown in Variant 2 of \cref{fig: abla}, compared to Variant 1, adding TARG reduces lip color changes in the eye-makeup task, leading to lower error in non-edited areas and higher identity-preservation scores across all datasets, as reported in \cref{tab: ablation}. 
This improvement results from projecting pixel masks into the attention domain, blocking non-ROI interactions, and reducing cross-region leakage, which stabilizes background geometry and prevents face rescaling.

Building on TARG, adding Cross-Modal Perception Guidance (CMPG) further enhances semantic alignment with the reference while maintaining the spatial stability achieved by TARG. As shown in Variant 3 of \cref{fig: abla}, compared to Variant 2, the eyeshadow transfer becomes more complete and stable. Makeup-similarity and identity scores improve, and the background remains stable, as reported in \cref{tab: ablation}. 
CMPG clarifies the preservation and transfer processes by aligning text and image cues during denoising, reducing content-boundary ambiguity, and further enhancing local style details and precision.

\section{Limitation}
Although MagicMakeup supports controllable full-face, eye, and lip makeup transfer, it does not yet provide finer-grained control over individual cosmetic components, such as blush or contour. In addition, performance may degrade under extreme side-view poses, where part of the makeup can be incompletely preserved, as illustrated in the fifth row of \cref{fig: Qua1}. This limitation is mainly due to the limited availability of real makeup data with extreme side-view poses, which remain rare in real-world datasets.

\section{Conclusion}
In this paper, we present MagicMakeup, a diffusion-transformer framework for high-fidelity, region-specific makeup transfer. Our system integrates Token-Aligned Region Gating to align pixel masks with attention and gate logits by region to prevent spillover, Cross-Modal Perception Guidance to align text and image features and make preservation/transfer decisions explicit, and an automated $1024 \times 1024$ makeup-removal pairing pipeline that supplies identity-consistent, region-labeled supervision. Additionally, we assemble a unified benchmark spanning synthetic and real data for standardized evaluation of regional controllability, makeup fidelity, and identity preservation. These components produce precise region-specific edits and achieve state-of-the-art results across diverse styles, races, and poses.

\section{Ethical Considerations and Dataset Release}
Our work uses real-face images for high-resolution facial makeup transfer, requiring careful consideration of privacy, consent, copyright, bias, and misuse risks. The benchmark is intended only for non-commercial academic research on cosmetic makeup transfer, not for identity recognition, face swapping, identity generation, or impersonation. To reduce these risks, we remove account identifiers, usernames, URLs, and unnecessary metadata and filter inappropriate, sensitive content during data curation. Restricted evaluation benchmark will be provided only through gated access under a Data Usage Agreement (DUA), which prohibits redistribution, identity-related use, and deceptive or malicious manipulation. We will also provide an opt-out/removal mechanism for individuals or content owners. We considered diversity in facial appearances and makeup styles during curation, while acknowledging that web-collected data may still reflect demographic, cultural, and style biases. Therefore, the benchmark should be used only within its intended research setting.




\bibliographystyle{splncs04}
\bibliography{main}


\newpage
\begin{subappendices}
\renewcommand{\thesection}{~\Alph{section}}
\clearpage
\setcounter{page}{1}

\section{Mask Visualization}

\begin{figure*}[!th]
\hsize=\textwidth 
\begin{center}
\includegraphics[width=0.95\linewidth]{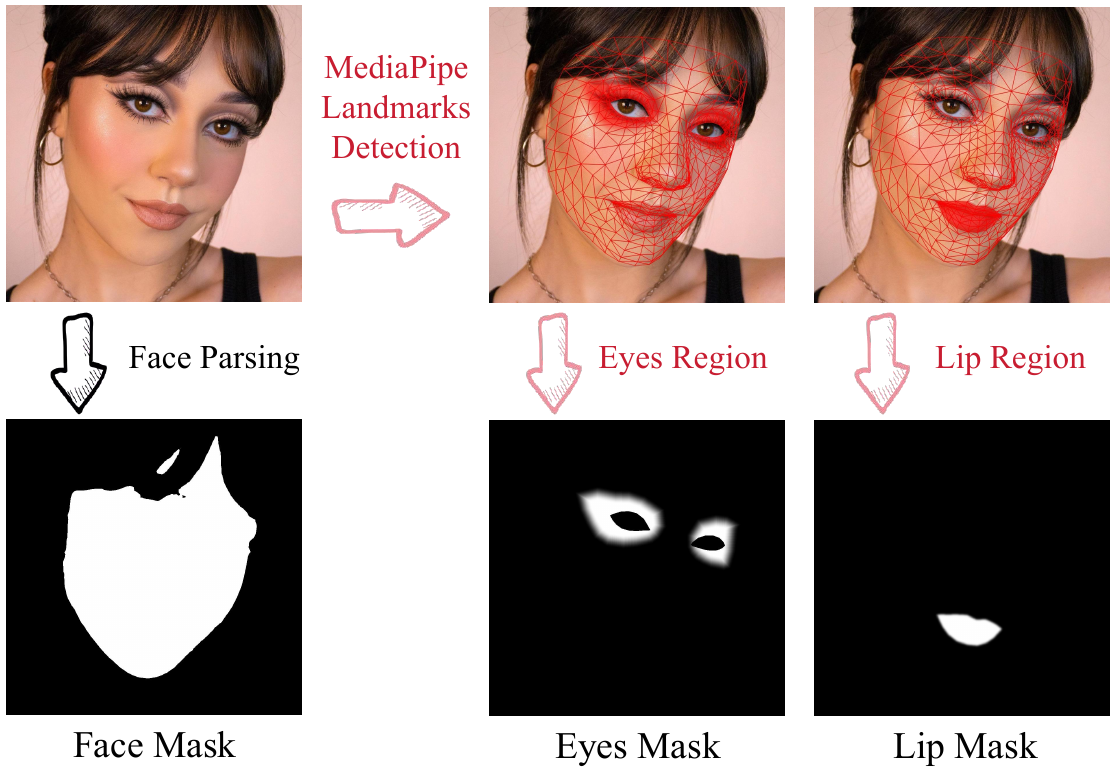}
\end{center}
\caption{\small \textbf{Visualization of mask acquisition.} The face mask is obtained by face parsing, and the eye and lip masks are derived from MediaPipe landmarks.}
\label{fig: mask}
\end{figure*}

\cref{fig: mask} provides the region masks used in our method. The full-face mask is obtained from face parsing \cite{{xie2021segformer}}, while the eye and lip masks are derived from predefined MediaPipe facial landmark \cite{kartynnik2019real} subsets. For the eyeshadow region, we use a soft mask derived from the eye area, as described in the main paper.

\section{Robustness to Mask Perturbations}

To examine whether our method heavily depends on precise masks, we evaluate its robustness under random mask perturbations, including small translations, rotations, and boundary blurring. As shown in \cref{tab: mask}, these perturbations cause only marginal performance changes: Face-ID decreases from 0.979 to 0.964, while DINO-I and CLIP-I remain almost unchanged.

These results suggest that the masks mainly serve as coarse spatial priors rather than the primary source of performance. The fine-grained control instead comes from CMPG, which further refines the preservation/transfer semantics beyond the spatial guidance provided by the masks. Even when the masks are imperfect, CMPG helps maintain stable identity preservation and coherent makeup transfer, which is consistent with the ablation results in \cref{fig: abla} in the main paper.

\begin{table}[t]
\centering
\caption{\textbf{Robustness to mask perturbations.} We randomly perturb the masks with small translations, rotations, and boundary blurring. The limited performance drop suggests that our method is robust to minor mask inaccuracies.}
\label{tab: mask}
\begin{tabular}{lccc}
\toprule
\textbf{Metric} & \textbf{Face-ID$\uparrow$} & \textbf{DINO-I$\uparrow$} & \textbf{CLIP-I$\uparrow$} \\
\midrule
Original  & 0.979 & 0.783 & 0.875 \\
Perturbed & 0.964 & 0.781 & 0.871 \\
\bottomrule
\end{tabular}
\end{table}

\section{MakeupHQ Bench}

\begin{figure*}[!ht]
\begin{center}
\includegraphics[width=0.95\linewidth]{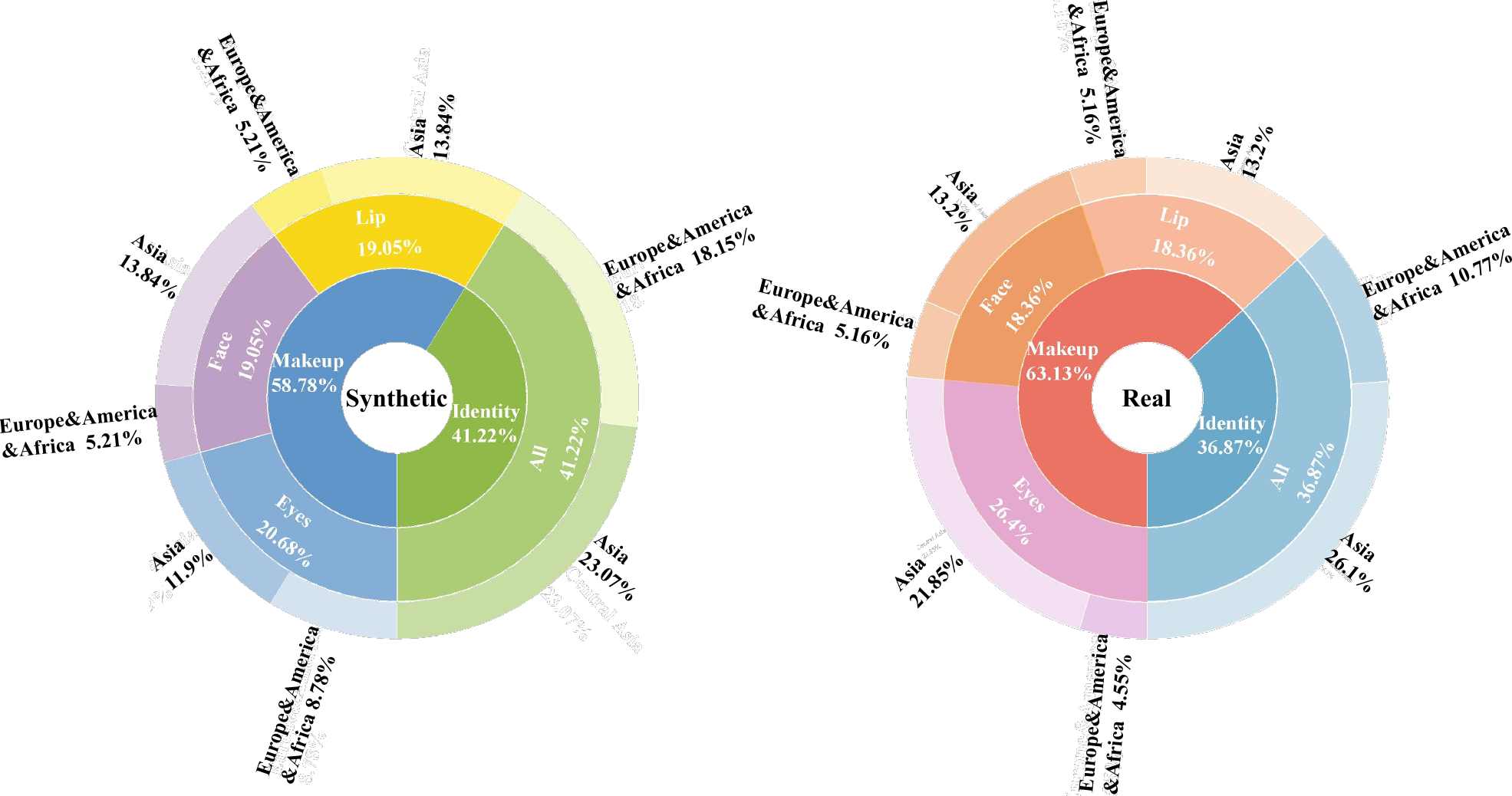}
\end{center}
\caption{\small
\textbf{Composition of the MakeupHQ Benchmark subsets.} Left: MakeupHQ-Synthetic subset; right: MakeupHQ-Real subset. For each subset, the inner ring separates source face cases from makeup reference cases, the middle ring further divides the makeup category into full-face, eyes, and lip edits, and the outer ring reports the distribution of races across major geographic groups.}
\label{fig: bench}
\end{figure*}

The structure of MakeupHQ Bench for region-specific evaluation is shown in \cref{fig: bench}. The inner ring shows the proportion of source face images and makeup reference images, while the middle ring categorizes the makeup references into full-face, eye, and lip edits, ensuring balanced representation across these three regions. The outer ring displays the distribution of subjects across major geographic groups, with Asia and Europe/America/Africa as the primary categories. Both the synthetic and real subsets feature a diverse mix of makeup styles, races, and poses, reflecting the varied styles, expressions, and viewpoints present in real-world makeup scenarios. This diverse and balanced composition ensures that the benchmark can support fine-grained, region-specific evaluation while maintaining broad representativeness across different cultural contexts.

\section{User Study}

We conduct a user study to compare different models quantitatively. We compare our method with four representative baselines selected for their strongest visual quality in our qualitative results. We randomly select 15 pairs from MakeupHQ-Real and 15 pairs from MakeupHQ-Synthetic, covering diverse makeup styles. For each pair, we generate results using the selected methods. We recruit 100 participants and ask them to select the most satisfactory result in terms of identity preservation and makeup fidelity. All results are presented simultaneously in a randomized order to ensure a fair comparison.

The results of the user study are shown in \cref{tab: userstudy}, where we report the percentage of participants who selected each method for real and synthetic images, as well as the overall mean selection rate. According to the study, it is evident that MagicMakeup achieves the highest average selection rate in both real and synthetic data, with a mean of 77.43\%.

\begin{table}[t]
\centering
\caption{\textbf{User study.} The ratio selected as best (\%) on MakeupHQ benchmark. Our method achieves the highest average score across both real and synthetic data, demonstrating superior performance in makeup-transfer.}
\setlength{\tabcolsep}{4pt}
\footnotesize
\resizebox{1\linewidth}{!}{
\begin{tabular}{l|cccccc}
\toprule
\textbf{Methods} & \textbf{CSD-MT} & \textbf{EleGANt} & \textbf{Stable-Makeup} & \textbf{Flux-Makeup} & \textbf{MagicMakeup} \\
\midrule
Real      & 2.4\%  & 1.47\%  & 3.8\%  & 5.67\% & 86.67\%  \\
Synthetic & 2.46\%  & 2.27\%  & 14.4\%  & 12.67\% & 68.2\%  \\
Average   & 2.43\%  & 1.87\%  & 9.1\%  & 9.17\% & 77.43\% \\
\bottomrule
\end{tabular}}
\label{tab: userstudy}
\end{table}

\begin{figure*}[!t]
\centering
\includegraphics[width=0.9\linewidth]{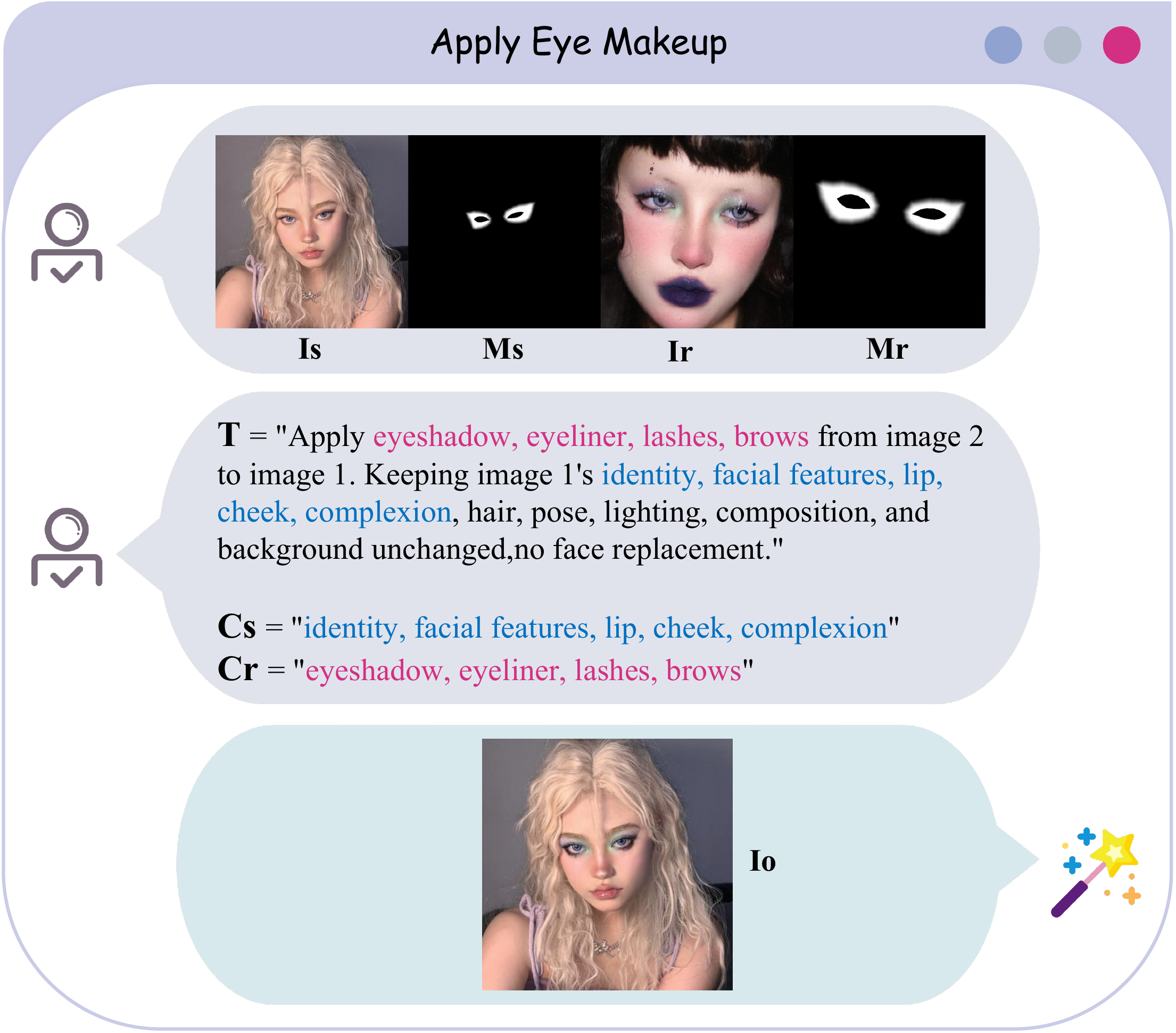}
\vspace{6pt}

\includegraphics[width=0.9\linewidth]{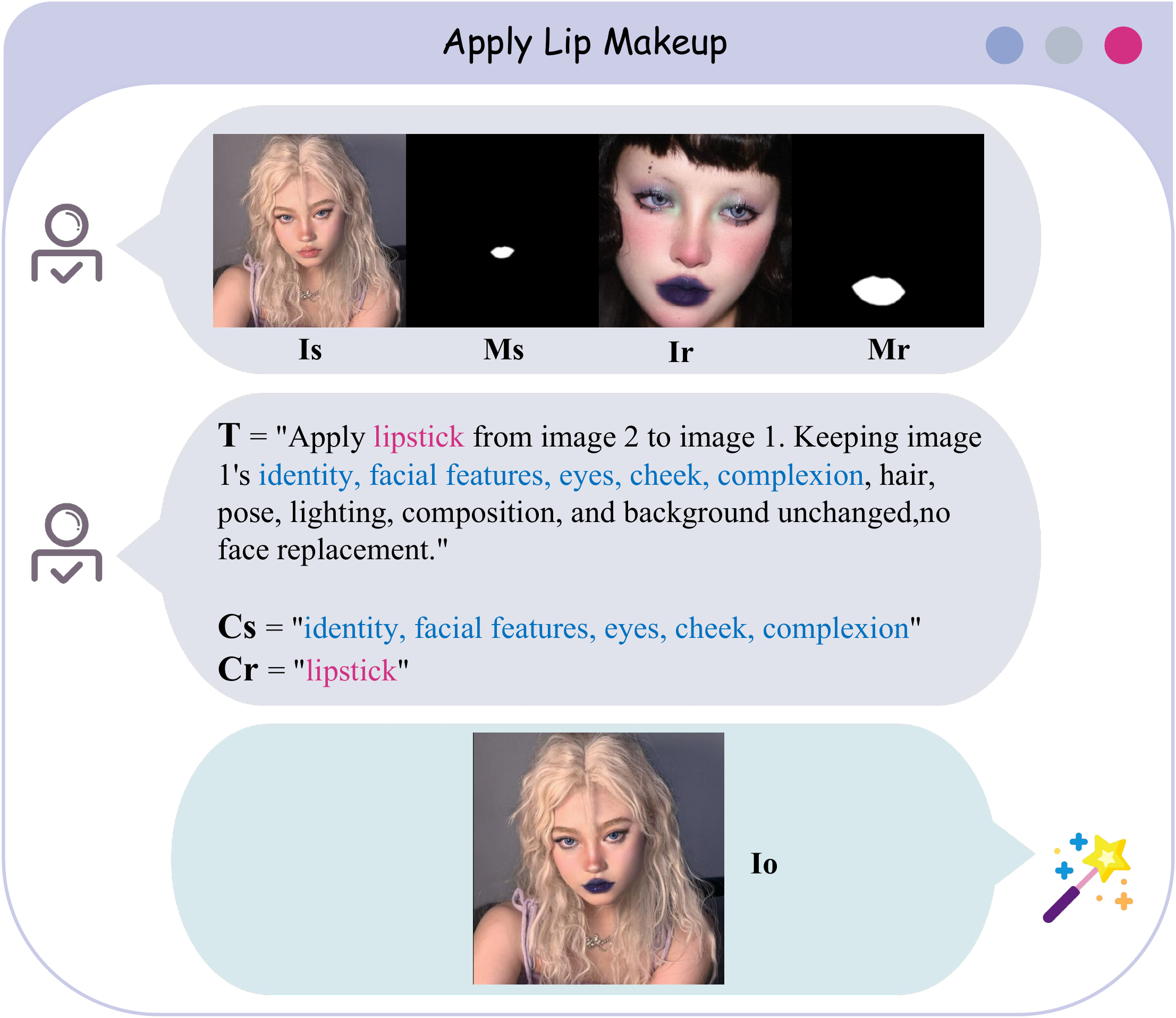}
\vspace{-6pt}

\caption{\small\textbf{Prompt templates for eye and lip makeup transfer.}}
\label{fig: prompt1}
\vspace{-10pt}
\end{figure*}

\section{Prompt Templates for Different Makeup Regions}

\begin{figure*}[t]
\hsize=\textwidth 
\begin{center}
\includegraphics[width=0.9\linewidth]{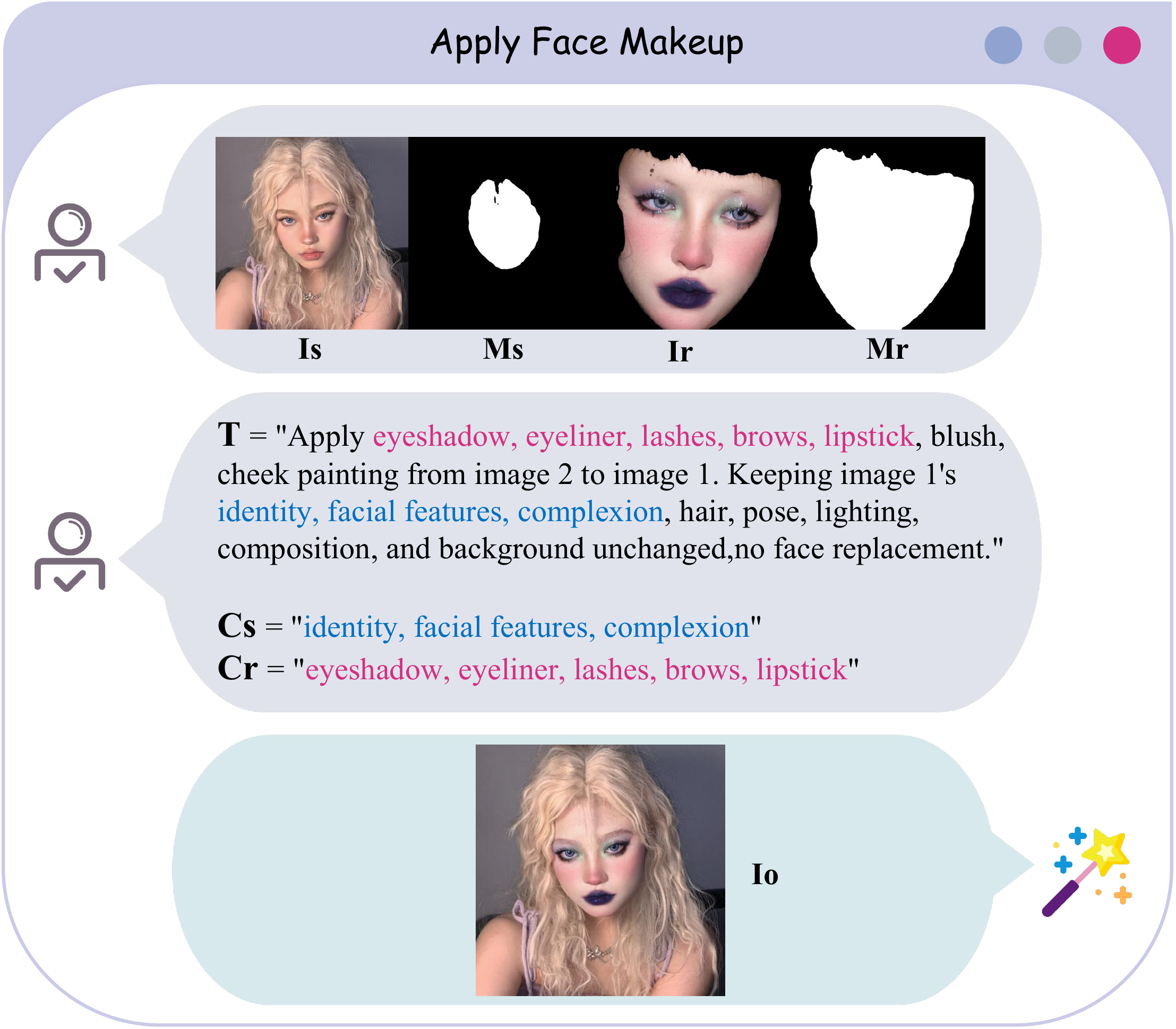}
\end{center}
\caption{\small
\textbf{Prompt template for full-face makeup transfer.}}
\label{fig: prompt3}
\end{figure*}

We provide prompt templates for the three controllable regions in \cref{fig: prompt1,fig: prompt3}. 
Each template specifies the source image $\mathbf{I_s}$ and its mask $\mathbf{M_s}$, the reference image $\mathbf{I_r}$ and its mask $\mathbf{M_r}$, the text prompt $\mathbf{T}$, the preservation concept $\mathbf{C_s}$ and the transfer concept $\mathbf{C_r}$ used by our method.

For full-face makeup transfer, the instruction covers the complete makeup attributes from the reference image, while the preservation concept emphasizes identity-related facial properties of the source image. For regional editing, the templates follow the same structure but restrict the transfer concept to the selected region. These examples illustrate how our method unifies full-face transfer and region-specific editing under the same prompting format.

\clearpage

\begin{figure}[!t]
\begin{center}
\includegraphics[width=0.95\linewidth]{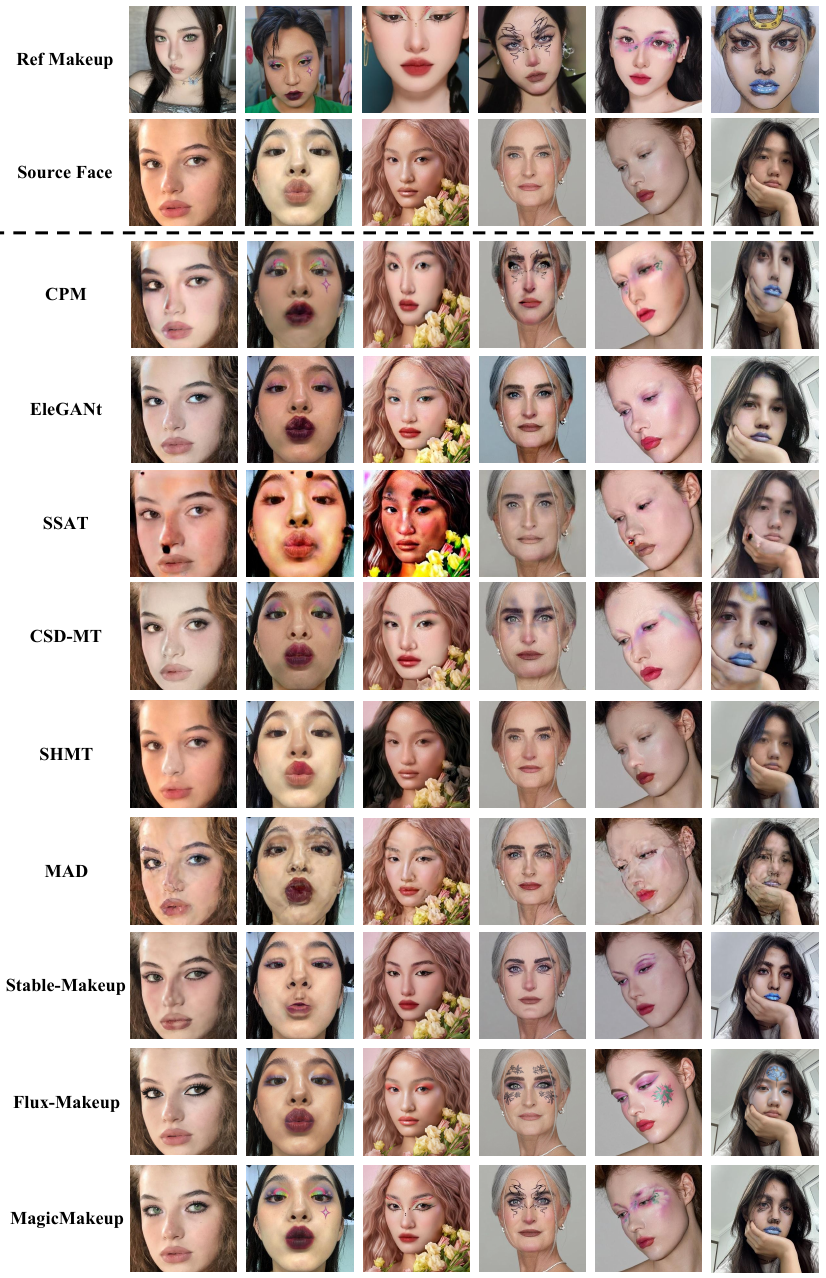}
\end{center}
\vspace{-4pt}
\caption{\small
\textbf{Comparison of our method with existing makeup-transfer approaches on the MakeupHQ benchmark.}
Our model excels in preserving identity while effectively transferring diverse makeup styles across both real and synthetic data.}
\vspace{-4pt}
\label{fig: Compall}
\end{figure}

\begin{figure}[!t]
\begin{center}
\includegraphics[width=\linewidth]{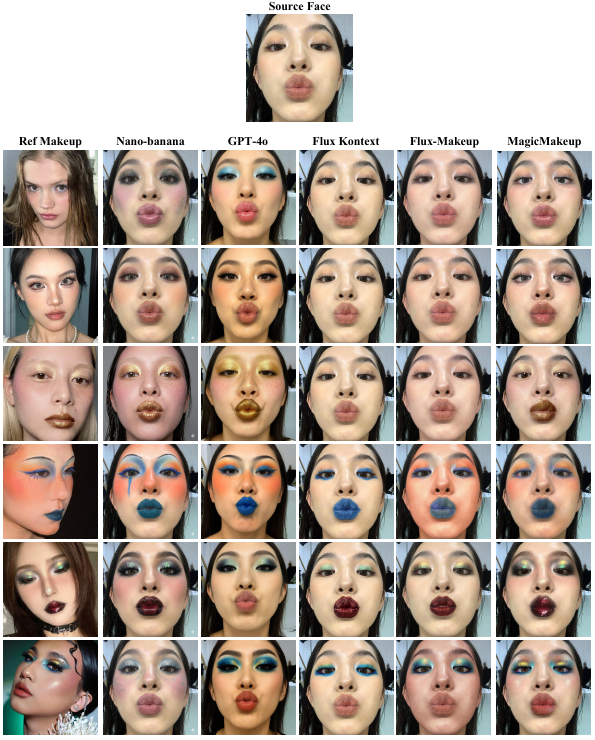}
\end{center}
\caption{\small
\textbf{Comparison between our method and general-purpose image-editing models (Nano-banana, GPT-4o, Flux Kontext) and the current SOTA makeup-transfer model Flux-Makeup.} The figure highlights that our method outperforms these general models in handling fine-grained makeup details and identity preservation, showing its superiority for makeup-transfer tasks.}
\label{fig: Compnano}
\end{figure}

\section{Makeup-Transfer Methods Comparison}

As \cref{fig: Compall} shows, we conduct an additional qualitative comparison between our method and existing open-source makeup-transfer methods on the MakeupHQ bench. This comparison highlights our model’s ability to preserve identity while transferring diverse makeup styles across both real and synthetic data.

\section{Image-editing Models Comparison}

We compare our method with image-editing models such as Nano-banana\cite{team2023gemini}, GPT-4o \cite{hurst2024gpt}, Flux Kontext \cite{labs2025flux}, and the current SOTA makeup-transfer model, Flux-Makeup \cite{zhu2025flux}. As shown in \cref{fig: Compnano}, these general image-editing models fail to effectively handle the fine-grained makeup details and identity preservation required for makeup transfer tasks. In contrast, our method is specifically designed for makeup transfer and outperforms these general models in maintaining both identity and makeup consistency, even when processing complex makeup styles and intricate facial features. Furthermore, our method surpasses Flux-Makeup in both makeup fidelity and identity preservation.

\clearpage

\section{Fine-Grained Performance Across Real-World Conditions}

\begin{figure*}[!t]
\hsize=\textwidth 
\begin{center}
\includegraphics[width=0.9\linewidth]{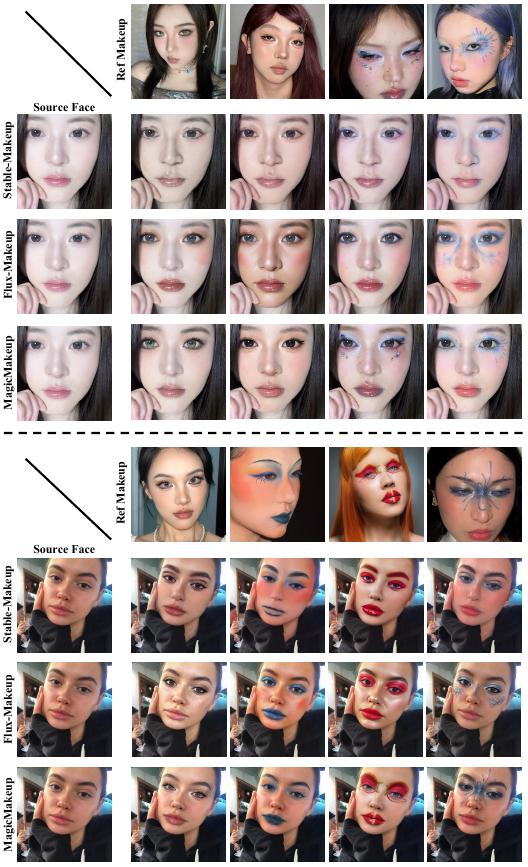}
\end{center}
\vspace{-10pt}
\caption{\small
\textbf{Performance under real-world variations.} Across large changes in pose, makeup style, and background, our method better preserves both the intended makeup and the source identity than Stable-Makeup and Flux-Makeup.}
\vspace{-10pt}
\label{fig: robust1}
\end{figure*}

\begin{figure*}[!th]
\hsize=\textwidth 
\begin{center}
\includegraphics[width=0.9\linewidth]{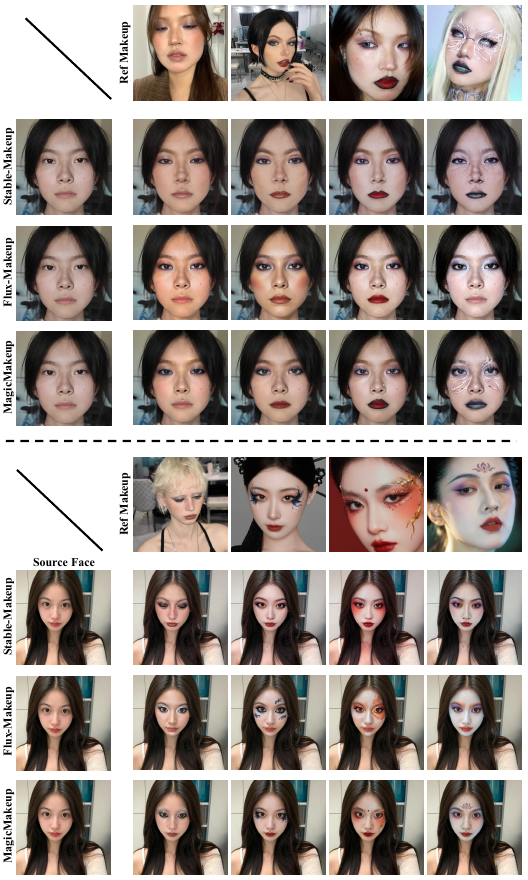}
\end{center}
\vspace{-10pt}
\caption{\small
\textbf{Performance under real-world variations.} Across large changes in pose, makeup style, and background, our method better preserves both the intended makeup and the source identity than Stable-Makeup and Flux-Makeup.}
\vspace{-10pt}
\label{fig: robust2}
\end{figure*}

In \cref{fig: robust1,fig: robust2}, we further compare our method with Stable-Makeup and Flux-Makeup under more challenging real-world variations. Using four source images paired with diverse reference styles, poses, and backgrounds, we observe that our method achieves more accurate makeup transfer while better preserving facial identity and overall visual consistency. Compared with the baselines, our results remain more stable under changes in pose, background, and makeup style, highlighting the advantage of our method in delivering both fine-grained controllability and robust real-world performance.

\begin{figure*}[!ht]
\begin{center}
\includegraphics[width=0.8\linewidth]{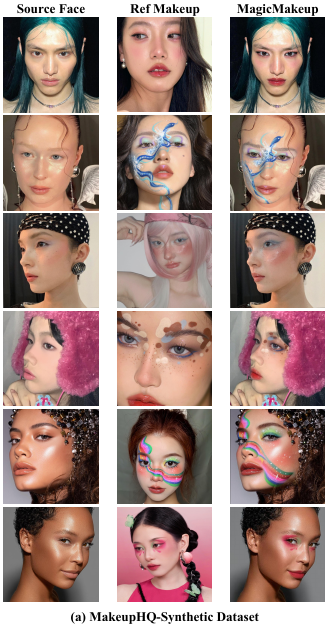}
\end{center}
\vspace{-4pt}
\caption{\small
\textbf{Fine-grained results on MakeupHQ-Synthetic.} Our method preserves identity while transferring diverse reference makeup styles and skin tones across synthetic source-reference pairs.} 
\vspace{-4pt}
\label{fig: data1}
\end{figure*}

\begin{figure*}[th]
\begin{center}
\includegraphics[width=0.8\linewidth]{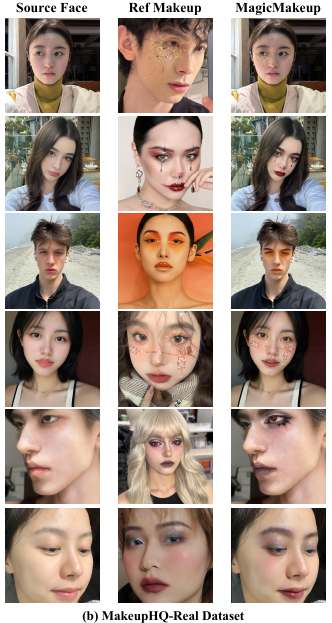}
\end{center}
\vspace{-4pt}
\caption{\small
\textbf{Fine-grained results on MakeupHQ-Real.} Our method generalizes to real-world sources and references with diverse identities, poses, and makeup styles.}
\vspace{-4pt}
\label{fig: data2}
\end{figure*}

In \cref{fig: data1,fig: data2}, we present additional results of our method on the two benchmarks, covering source faces with diverse facial characteristics and skin tones, as well as reference images with different makeup styles. The results show that our method generalizes well across substantial variations in appearance and reference makeup, while maintaining stable and fine-grained transfer quality. In particular, it produces coherent makeup effects across different facial structures, expressions, and poses, demonstrating strong adaptability in realistic scenarios.

\end{subappendices}

\end{document}